\documentclass[letterpaper]{article} 
\usepackage{aaai2027}  
\usepackage[hyphens]{url}  
\usepackage{graphicx} 
\usepackage{natbib}  
\usepackage{caption} 
\usepackage[ruled,vlined,linesnumbered]{algorithm2e}

\usepackage{amsmath} 
\usepackage{ntheorem}

\usepackage{float}

\usepackage{listings}
\DeclareCaptionStyle{ruled}{labelfont=normalfont,labelsep=colon,strut=off} 
\usepackage{booktabs}

\usepackage{amssymb}

\title{\textsc{GRALS}: GCN-Guided Redundancy-Aware Local Search for Minimum Vertex Cover}
\author{
    Chanjuan Liu\textsuperscript{\rm 1},
    Qiqi Bao\textsuperscript{\rm 1},
    Yu Zhang\textsuperscript{\rm 2},
    Yaochu Jin\textsuperscript{\rm 3},
    Enqiang Zhu\textsuperscript{\rm 4}\corresponding
}
\affiliations{
\textsuperscript{\rm 1}School of Computer Science and Technology, Dalian University of Technology, Dalian 116024, China\\
    \textsuperscript{\rm 2}School of Computer Science, Peking University, Beijing 100091, China\\
    \textsuperscript{\rm 3}School of Engineering, Westlake University, Hangzhou 310024, China\\
    
    \textsuperscript{\rm 4}Institute of Computing Science and Technology, Guangzhou University, Guangzhou 510006, China
}

\begin{document}

\maketitle

 \begin{abstract}

The minimum vertex cover (MVC) problem seeks to identify the smallest set of vertices that cover all edges in an undirected graph. As a fundamental NP-hard combinatorial optimization problem, MVC has been widely studied due to its applications in network analysis and system design. For large-scale instances, local search heuristics are among the most effective approaches, offering a practical trade-off between solution quality and computational efficiency. Most existing high-performance local search algorithms adopt a break-and-repair framework, where the repair process mainly restores feasibility but provides limited opportunities for enhancing the current solution's structure. We introduce \textsc{GRALS}, a local search framework that integrates vertex probability priors learned from a graph convolutional network with an expansion–revelation–elimination (ERE) operator. The learned priors guide the search toward promising regions, while the ERE operator dynamically identifies and eliminates redundant vertices to further enhance solution quality. Experiments conducted on 346 benchmark instances demonstrate that \textsc{GRALS} achieves the best-known solution for 335 instances (96.82\%), compared to 302 instances (87.28\%) achieved by the strongest baseline. The advantage of \textsc{GRALS} is maintained across different time limits and is particularly evident on large-scale graphs with millions of vertices. 
 \end{abstract}

\section{Introduction}
The minimum vertex cover (MVC) problem is to find a vertex subset \(C\) of minimum cardinality in a graph such that every edge is incident with at least one vertex in \(C\). As one of Karp's 21 NP-hard problems~\cite{Karp1972}, MVC has found broad applications in various fields, including communication modeling \cite{dagdeviren2023metaheuristic}, wireless sensor systems \cite{yigit2021breadth}, and large-scale network analysis  \cite{Kavalci2014}. 


Existing MVC algorithms can be classified into three main categories: exact, approximation, and heuristic algorithms. Exact algorithms, often based on branch-and-bound techniques, guarantee optimal solutions but typically suffer from poor scalability due to the exponential worst-case complexity of the problem~ \cite{chen2010improved,Akiba2016}. Approximation algorithms provide polynomial-time guarantees; however, inherent inapproximability bounds restrict their worst-case performance ratios~\cite{Dinur2005,Vazirani2001}. Consequently, for practical applications involving large-scale graphs, local search heuristics have emerged as the preferred method, effectively balancing efficiency with solution quality~\cite{cai2015balance,Gu2021,Zhang2023,Sun2024}.

Modern high-performance MVC local search algorithms primarily adopt a \emph{break-and-repair} paradigm. This method starts with a feasible solution and iteratively perturbs it (breaks) before restoring its feasibility (repairs) to find smaller vertex covers. Within this framework, many techniques has emerged, including configuration checking and edge weighting~\cite{cai2011local}, two-weighting and balance-oriented heuristics designed for massive  graphs~\cite{Cai2015,cai2015balance}, noisy diversification strategies~\cite{MaFSLS16}, the integration of construct-local-search-preprocess approaches~\cite{CaiLL17}, automatic configuration methods~\cite{Luo2019}, edge-age-based strategies~\cite{quan2021local,Gu2021}, and rule-based inference exchange techniques~\cite{Sun2024}. These advancements have enhanced practical performance, particularly on large sparse graphs.

Despite these advances, two limitations remain. First, current break-and-repair algorithms address redundant vertices only when they are passively encountered, rather than proactively creating redundancy to enable further reductions. This reactive approach may miss opportunities for deeper compression. Second, the initialization process relies on local heuristics (e.g., degree metrics \cite{Cai2013}, greedy edge coverage \cite{Zhang2023}, or rule-based strategies~\cite{Sun2024}) that do not sufficiently capture multi-hop structural information required to guide the subsequent search toward optimization.



To address these limitations, we introduce GRALS, a new break-and-repair local search for MVC. GRALS consists of three key components: a GCN-guided initialization that converts learned vertex priors into edge ownership information, leading to high-quality initial solutions; an expansion–revelation–elimination (ERE) operator that actively generates and leverages redundancy to enhance the solution improvement process; and a proactive redundancy recovery (PRR) mechanism that, during periods of search stagnation, guides repair decisions towards areas with greater potential for releasing redundancy. By maintaining zero-loss and critical-vertex states throughout the search, GRALS refines the traditional break-and-repair framework by integrating explicit redundancy information, achieving more efficient solution reductions. In evaluations involving 346 benchmark instances, GRALS achieved the best-known solution in 335 cases (96.82\%), surpassing the strongest baseline, which solved 302 cases (87.28\%). This advantage is particularly impressive on large-scale sparse graphs.

The remainder of this paper is organized as follows: Section 2 provides the necessary preliminaries; Section 3 details GRALS and its components; Section 4 presents the experimental results; and Section 5 concludes the paper.

\section{Preliminaries}\label{section-2}
This section introduces the fundamental notation and structural concepts used throughout the paper. 


  \subsection{Definitions and Notations}
All graphs considered are simple undirected graphs. Let $G=(V, E)$ be a graph with a \textit{vertex set} $V$ and an \textit{edge set} $E$. For a vertex $v\in V$, its \textit{open neighborhood} is defined as
  $
  N(v)=\{u \mid (u\in V) \land (uv\in E)\}
  $, its \textit{closed neighborhood} as $N[v] =N(v)\cup \{v\}$, and its \textit{degree} as $d(v)=|N(v)|$. Given a subset $C\subseteq V$, an edge $uv \in E$ is \textit{covered} by $C$ if $\{u,v\} \cap C \neq \emptyset$, and is \emph{double-covered} by $C$  if $\{u,v\} \subseteq C$. If every edge in $E$ is covered by $C$, then $C$ is called a \emph{vertex cover} (VC) of $G$, with the vertices in \( C \) termed the \textit{cover vertices}.   The minimum vertex cover (MVC) problem is to identify a VC of minimum cardinality.

To facilitate the subsequent presentation, we introduce the following additional notions. Let $C\subseteq V$ be an arbitrary candidate vertex subset in a graph $G=(V, E)$.  For a vertex $v\in C$, we define its \textit{deletion loss} as
  $
  \ell_C(v)
  =
  \bigl|\{u \mid (u\in N(v)) \land (u\notin C)\}\bigr|,
  $
which counts the number of edges that would become uncovered if $v$ were removed from $C$. A vertex $v\in C$ is called \emph{zero-loss} if $\ell_C(v)=0$, and \emph{critical} if $\ell_C(v)=1$. We denote the sets of zero-loss and critical vertices by  $Z(C)=\{v\in C\mid \ell_C(v)=0\}$ and 
  $Q(C)=\{v\in C\mid \ell_C(v)=1\}$, respectively. Removing a zero-loss vertex introduces no new uncovered edges; in particular, if $C$ is a VC, then $C\setminus\{v\}$ remains a VC. A critical vertex becomes zero-loss exactly when its unique neighbor outside $C$ is inserted into the candidate set $C$. For a vertex $x \in V\setminus C$, let \(S_C(x)=N(x)\cap Q(C)\) and $R_C(x) = |S_C(x)|$. 
 For a vertex $v \in V$, its $dscore$ is defined as: if $v\notin C$, then $dscore(v)$ equals the number of uncovered edges incident with $v$; if $v\in C$,  $dscore(v)=-\ell_C(v)$. We define the \textit{cover-neighbor count} of $v$ as
$
cn_C(v) = |N(v) \cap C|.
$
When \( v \notin C \), adding \( v \) to $C$ results in exactly \( cn_C(v) \) incident edges being double-covered. We denote the set of uncovered edges by \( C \) as 
$
U(C) = \{uv \in E \mid u \notin C, \ v \notin C\}.
$
Thus, \( C \) qualifies as a VC if and only if \( U(C) = \varnothing \).


  \subsection{Break-and-Repair Local Search}
The break-and-repair local search maintains a current candidate set \(C\), which is initialized from a feasible solution. In each iteration, a selected subset of vertices is removed from \(C\), potentially leaving some edges uncovered and thus making the solution infeasible. Then a repair step adds vertices back until all edges are covered again, restoring feasibility. 

  \subsection{Graph Convolutional Network}

  
The graph convolutional network (GCN) takes the adjacency matrix \(A \in \{0,1\}^{|V|\times |V|}\) of a graph \(G\) and the node-feature matrix \(X \in \mathbb{R}^{|V|\times f}\) as input. Following the standard formulation~\cite{KipfWelling2017}, it propagates information through successive layers using the update rule:
\[
H^{(\ell+1)} = \sigma\!\left( \widetilde{D}^{-\frac{1}{2}} \widetilde{A} \widetilde{D}^{-\frac{1}{2}} H^{(\ell)} W^{(\ell)} \right),
\]
where \(H^{(0)}=X\), \(\widetilde{A}=A+I\) adds self-loops, \(\widetilde{D}\) is the degree matrix of \(\widetilde{A}\), \(W^{(\ell)}\) is the trainable weight matrix at layer \(\ell\), and \(\sigma(\cdot)\) denotes a nonlinear activation function. This message-passing mechanism enables each vertex to gradually aggregate features from its multi-hop neighborhood, effectively merging local node attributes with a broader structural context. After being trained on reference VCs, the pretrained GCN generates a score \(\textit{prob}(v) \in [0,1]\) for every vertex \(v\), which indicates the learned likelihood of that vertex being part of a VC. These scores are utilized only as structural priors during the initialization and are neither updated nor queried during the subsequent search process.

 \section{Method}\label{section-3}

\textsc{GRALS} consists of three main components: GCN-based initialization, break-and-repair search, and redundancy optimization. The initialization leverages GCN probabilities to generate an effective starting VC $C$. The search then iteratively removes vertices from $C$, and when the current candidate set becomes infeasible, a minimal set of vertices is added back to restore feasibility. Two mechanisms are employed to address redundancy: the \emph{expansion-revelation-elimination} (ERE) operator, which adds and removes vertices to expose redundancy, and the \textsc{proactive redundancy recovery} (PRR) strategy, activated during search stagnation, directs repair efforts toward areas with greater potential for redundancy.  A light perturbation is applied only after prolonged stagnation to diversify the search.

\subsection{GCN-Guided Initialization}



For an instance graph $G = (V, E)$, each vertex $v$ is characterized by four features: a constant feature valued at 1\cite{liu2023accurately},  the vertex degree $d(v)$ \cite{chen2012identifying}, a normalized edge-density feature $|E|/|V|^2$\cite{pu2022integrated}, and a normalized average-neighbor-degree feature $\frac{1}{d(v)|N(v)|}\sum_{u\in N(v)} d(u)$  related to the friendship index \cite{pal2019study}. We employ a 2-layer GCN with a hidden dimension of 64, ReLU activation, and a sigmoid output \(prob(v)\). For smaller graphs, labels are derived from optimal VCs obtained by the exact solver EMVC \cite{math7070603}, while for larger graphs, high-quality VCs generated by local search algorithms proposed in \cite{Zhang2023,Sun2024} within a 1000-second limit serve as pseudo-labels, allowing the model to be trained using binary cross-entropy. Since the GCN's primary objective is to rank candidate vertices rather than predict a VC, this mixed supervision approach is adequate: Smaller graphs offer reliable exact patterns, while larger graphs expose the model to structures closer to the test instances. Training occurs offline and is not included in the reported runtime; test-time inference is conducted on a GPU and is counted within the 1000-second budget, while initialization and local search run on CPU.

%

Let \(S\) denote the set of vertices already selected into the current partial VC. The \textit{residual graph} is defined as \(G_R = (V_R, E_R)\), where \(V_R = V \setminus S\) and \(E_R = \{\, uv \in E \mid u, v \in V_R \,\}\). For each vertex \(v \in V_R\), we define its \textit{residual degree} in \(G_R\) as \(d_R(v) = d_{G_R}(v)\) and define its \textit{priority score} $pscore(v)$ as \(pscore(v) =  d_R(v) + \alpha(v)\), where 
$\alpha(v)=\bigl| \{\, x \in (N(v) \cap V_R) \mid prob(v) > prob[x]  \,\} \bigr|$
and $prob(v)$ is the probability derived from the GCN. This formulation integrates the residual degree with a GCN-derived bias that favors vertices with a higher likelihood of inclusion, guiding the search toward promising candidates for the next selection.

Three low-degree exact reduction rules~\cite{Fan2015,Chen2001} are applied iteratively before each heuristic selection step until no such reductions are available.

\textbf{Reduction Rule 1.} \emph{Let $v \in V_R$ be a one-degree vertex and $u$ its unique neighbor. We add $u$ to $C$ and delete $v$ from $G_R$.}

\textbf{Reduction Rule 2.} \emph{Let $v \in V_R$ be a two-degree vertex whose two neighbors $u$ and $w$ are adjacent. We add both $u$ and $w$ to $C$ and delete $v$ from $G_R$.}


\textbf{Reduction Rule 3.} \emph{Let $u$ and $v$ be two degree-two vertices that share the same two non-adjacent neighbors, $n_1$ and $n_2$. We add $n_1$ and $n_2$ to $C$ and remove both $u$ and $v$ from $G_R$.}

\setlength{\textfloatsep}{0pt plus 0pt minus 0pt}
\setlength{\floatsep}{0pt plus 0pt minus 0pt}

{

\setlength{\textfloatsep}{3pt}
\setlength{\floatsep}{3pt}
\begin{algorithm}[!t]
\scriptsize
\SetAlgoSkip{smallskip}
\SetNlSkip{1pt}
\setlength{\algomargin}{0em}
\setlength{\rightskip}{-0.6em}
\SetInd{0.2em}{0.45em}
\setlength{\algomargin}{0em}
\caption{\textsc{GCNInitVC}}
\label{alg:initial-solution}
\KwIn{A graph $G=(V,E)$}
\KwOut{A VC $C$}

Compute $prob(v)$ for each $v\in V$ via GCN\;
$C\leftarrow\emptyset$\;
$pscore(v)\leftarrow d_R(v)$ for all $v\in V$\;

\ForEach{$uv\in E$ with $prob(u)\neq prob(v)$}{
    $w\leftarrow\arg\max_{x\in\{u,v\}}prob(x)$\;
    $pscore(w)\leftarrow pscore(w)+1$\;
}

$G_R(V_R, E_R) \leftarrow G$\;

\While{$E_R\neq\emptyset$}{
   $P\leftarrow\{v\in V_R\mid d_R(v)\in\{1,2\}\land v\text{ satisfies Rule 1, 2, or 3}\}$\;
    \eIf{$P\neq\emptyset$}{
        $x\leftarrow$ select a vertex from $P$\;
        $S\leftarrow$ the vertices selected into $C$ by Reduction Rule 1, 2, or 3\;
    }{
        $e=v_1v_2\leftarrow$ edge of $G_R$ with maximum residual degree difference between endpoints\;
        $v\leftarrow\arg\max_{x\in\{v_1,v_2\}}pscore(x)$\;
        $S\leftarrow\{v\}$\;
    }
    $C\leftarrow C\cup S$\;
    $V_R \leftarrow V_R \setminus S$\;
    $E_R \leftarrow E_R \setminus \{e \in E_R \mid e \text{ is covered by }S\}$\;
    Update $pscore$ on the updated graph $G_R=(V_R, E_R)$\;
}

Remove redundant vertices from $C$\;
\Return{$C$}\;

\end{algorithm}
}

Algorithm~\ref{alg:initial-solution} outlines the initialization procedure. It begins by calculating the GCN probability $prob(v)$ for each vertex and initializing the VC $C$, as well as the priority score $\mathit{pscore}$ (lines 1--6). 
After setting the residual graph as $G_R=G$ (line 7), the
algorithm repeatedly processes $G_R$ until $E_R$ becomes empty
(lines 8--20). In each iteration, it starts by identifying vertices with degree one or two. If such vertices exist, it selects one and applies an appropriate reduction rule on it (lines 9--12). If not, it selects an edge in $G_R$ whose endpoints, say $u$ and $v$, have the maximum
$|d_R(u)-d_R(v)|$ difference and chooses the endpoint with the higher $\mathit{pscore}$ (lines 13--16). The selected vertices are added to $C$ and deleted from the residual graph, and priority scores are updated accordingly (lines 17--20). Finally, it removes redundant vertices and returns the resulting VC (lines 21--22). 
Following InfVC~\cite{Sun2024}, we employ CPLEX 22.10 to exactly solve each component with fewer than 300 vertices.

Both the computation of GCN probabilities~\cite{chiang2019cluster} and $pscore$ initialization require $O(|V|+|E|)$ time. The main loop maintains a max-heap based on the degree difference of the endpoints of each edge, with each heap operation costing $O(\log |E|)$. When a vertex is removed, it updates the heap for all incident edges, resulting in $O(|V||E|)$ updates over at most $|V|$ removals. Consequently, the loop runs in $O(|V||E|\log|E|)$. CPLEX is called only on components of constant size (fewer than 300 vertices), which adds negligible constant time. Therefore, Algorithm~\ref{alg:initial-solution} has an overall complexity of $O(|V||E|\log|E|)$.


\subsection{\textsc{GRALS} Algorithm}

Given an input instance \(G = (V, E)\), \textsc{GRALS} conducts a local search over a candidate set \(C \subseteq V\), which may either be feasible (a VC) or infeasible (a partial VC). The algorithm operates in three modes: \textsc{Normal}, \textsc{PRR}, and \textsc{Repair}. 
In the \textsc{Normal} mode, it implements standard remove-add operations. When the search stagnates, it transitions to \textsc{PRR}, which replaces the typical addition with a redundancy-aware procedure that leverages the structural characteristics of the current candidate vertex set. The \textsc{Repair} mode is activated following a perturbation, aiming to help the search quickly regain a feasible solution from an infeasible one.
 Before detailing the algorithm, we define several supporting routines:

\textsc{FoldOpt}: Let \(x \in V \setminus C\) be a vertex for which there are \(u, w \in S_C(x)\) such that \(uw \notin E\). \textsc{FoldOpt} adds  \(x\) to \(C\) and remove \(u\) and \(w\) from \(C\). It minimizes the size of the candidate vertex set, as established in previous works~\cite{andrade2012fast, Sun2024}.

\textsc{SelectRemove}: If \(Z(C)\neq\emptyset\), \textsc{SelectRemove} selects a vertex from \(Z(C)\); otherwise, it chooses a vertex from $C$ with the highest \(\mathit{dscore}\).

\textsc{NormalAdd}: An uncovered edge is randomly selected, followed by the application of a taboo strategy to its endpoints. A vertex \(x\) is considered \textit{taboo} if none of its neighbors have been added to or removed from the solution since the last operation involving \(x\). We represent the state of \(x\) as \(\mathit{conf\_change}(x)=0\) for taboo and \(\mathit{conf\_change}(x)=1\) for non-taboo. If only one endpoint is taboo, the non-taboo endpoint is chosen. If both endpoints are non-taboo, we prefer the vertex with the higher \(dscore\), breaking any ties by selecting the vertex with the smaller \(\mathit{time\_stamp}\), which records the iteration number when vertex \(v\) was last modified.

\textsc{RepairAdd}: It first selects \(\sqrt{|U(C)|}\) uncovered edges  randomly. Among the endpoints of the chosen edges, a non-taboo vertex with the highest \(\mathit{dscore}\) is then selected.

{
\SetAlgoSkip{}
\begin{algorithm}[!t]
\scriptsize

\SetNlSkip{1pt}
\setlength{\algomargin}{0em}
\setlength{\rightskip}{-0.6em}
\SetInd{0.2em}{0.45em}
\caption{\textsc{GRALS}}
\label{alg:search}

\KwIn{A graph $G=(V,E)$, cutoff time $t$, hyper-parameters
$\theta_{\mathrm{Pert}}$, $s_{\mathrm{PRR}}$,
$\theta_{\mathrm{PRR}}$, $\theta_{\mathrm{ERE}}$, $k$, $d$}
\KwOut{A VC $C^*$ of $G$}

$C\leftarrow\textsc{GCNInitVC}(G)$\;
$C^*\leftarrow C$;
$\mathit{search\_mode}\leftarrow\textsc{Normal}$;
$\mathit{unimp\_num}\leftarrow0$;
$\mathit{q\_gain}\leftarrow0$;
$\mathit{prr\_steps}\leftarrow0$;
$\mathit{no\_gain\_steps}\leftarrow0$\;
Initialize $\ell_C(\cdot)$, $Z(C)$, $Q(C)$,
$R_C(\cdot)$, and $cn_C(\cdot)$\;

\While{elapsed time $<t$}{
    \If{$U(C)=\emptyset$}{
        \While{
            $(\exists x\in V\setminus C)
            \land(\exists u,w\in S_C(x))
            \text{ s.t. }u\neq w\land uw\notin E$
        }{
            $C\leftarrow(C\cup\{x\})\setminus\{u,w\}$\;
        }

        \If{$|C|<|C^*|$}{
            $C^*\leftarrow C$;
            $\mathit{unimp\_num}\leftarrow0$;
            $prr\_steps\leftarrow0$;
            $\mathit{no\_gain\_steps}\leftarrow0$;
            $\mathit{search\_mode}\leftarrow\textsc{Normal}$\;
        }

        \eIf{$Z(C)\neq\emptyset$}{
            $z\leftarrow$ a vertex in $Z(C)$;
            $C\leftarrow C\setminus\{z\}$\;
        }{
            $r\leftarrow
            \arg\max_{v\in C}\mathit{dscore}(v)$;
            $C\leftarrow C\setminus\{r\}$\;
        }

        \textbf{continue}\;
    }

    \If{$\mathit{unimp\_num}>\theta_{\mathrm{Pert}}$}{
        $\mathit{uc\_num}\leftarrow|U(C)|$\;
        $S\leftarrow$ a random $d$-subset of $C$\;
        $C\leftarrow C\setminus S$\;
        $C\leftarrow C\cup
        \{\text{$d$ highest-degree vertices in }V\setminus C\}$\;
        $\mathit{search\_mode}\leftarrow\textsc{Repair}$;
        $\mathit{unimp\_num}\leftarrow0$;
        $\mathit{prr\_steps}\leftarrow0$;
        $\mathit{no\_gain\_steps}\leftarrow0$\;
    }

    \If{
        $\mathit{search\_mode}=\textsc{Repair}
        \land\mathit{uc\_num}\geq|U(C)|$
    }{
        $\mathit{search\_mode}\leftarrow\textsc{Normal}$\;
    }

    \If{
        $\mathit{search\_mode}=\textsc{PRR}
        \land
        \bigl(
        \mathit{prr\_steps}\geq s_{\mathrm{PRR}}
        \lor
        \mathit{no\_gain\_steps}\geq100
        \bigr)$
    }{
        $\mathit{search\_mode}\leftarrow\textsc{Normal}$\;
    }

    \If{$search\_mode=\textsc{Normal}
    \land unimp\_num>\theta_{\mathrm{PRR}}
    \land prr\_steps=0$}{
        $search\_mode\leftarrow\textsc{PRR}$\;
    }

    $r\leftarrow\textsc{SelectRemove}(C)$;
    $C\leftarrow C\setminus\{r\}$;

    \eIf{$\mathit{search\_mode}=\textsc{PRR}$}{
        $x\leftarrow\textsc{PRRSelect}(C,r,k), prr\_steps\leftarrow prr\_steps+1$\;
    }{
        \eIf{$\mathit{search\_mode}=\textsc{Normal}$}{
            $x\leftarrow\textsc{NormalAdd}(C)$\;
        }{
            $x\leftarrow\textsc{RepairAdd}(C)$\;
        }
    }

    $\mathit{q\_gain}\leftarrow
    \mathit{q\_gain}
    +\bigl|
        \bigl(Q(C\cup\{x\})\setminus Q(C)\bigr)
        \setminus\{x\}
    \bigr|$\;
    $C\leftarrow C\cup\{x\}$\;

    \If{
        $\mathit{search\_mode}=\textsc{Normal}
        \land Z(C)=\emptyset
        \land\mathit{q\_gain}\geq\theta_{\mathrm{ERE}}$
    }{
        $C\leftarrow
        \textsc{ERESelect}(C,\theta_{\mathrm{ERE}})$;
        $\mathit{q\_gain}\leftarrow
        \mathit{q\_gain}-\theta_{\mathrm{ERE}}$\;
    }

    $\mathit{unimp\_num}\leftarrow
    \mathit{unimp\_num}+1$\;
}

\Return{$C^*$}\;
\end{algorithm}
}

Algorithm~\ref{alg:search} outlines the proposed \textsc{GRALS}. 
It first generates an initial candidate set $C$ using GCNInitVC, sets the search mode to \textsc{Normal}, and clears the counters $\mathit{unimp\_num}$ and $\mathit{q\_gain}$. 
It also computes the auxiliary functions $\ell_C(\cdot)$, $Z(C)$, $Q(C)$, $R_C(\cdot)$, and $cn_C(\cdot)$ for $C$ (lines~1--3). 
The main loop runs until the time budget expires (line~4). If $C$ is feasible, the algorithm reduces it by repeatedly applying \textsc{FoldOpt}. When this reduction yields a smaller VC, it updates the current best solution $C^*$ and resets $\mathit{unimp\_num}$ (lines~5--9). 
Then it removes one vertex from $C$ to remove redundancy or break feasibility: preferentially a zero-loss vertex from $Z(C)$, or else the vertex $r\in C$ with the largest $\mathit{dscore}(r)$ (lines~10--14). The search adaptively changes its modes through the logic in lines~15--26:
If the search in \textsc{Normal} mode has not improved for too long time ($\mathit{unimp\_num} > \theta_{\mathrm{Pert}}$), the algorithm applies a random perturbation: It records $uc\_num = |U(C)|$, removes a random $d$-subset from $C$, and adds the $d$ highest-degree vertices in $V\setminus C$ to $C$. 
The mode is forced to \textsc{Repair}, and $\mathit{unimp\_num}$ is reset to zero (lines 15-20). The \textsc{Repair} mode terminates and reverts to \textsc{Normal} when $|U(C)| \le uc\_num$ (lines 21-22). The \textsc{PRR} mode stops when the maximum number of iterations $s_{\mathrm{PRR}}$ that a single PRR can perform is reached, or the number of consecutive iterations without removing any redundant vertex ($\mathit{no\_gain\_steps}$) reaches 100 (lines 23-24). 
These criteria prevent the algorithm from wasting time on stalled search regions, and allow it to fall back to the broader \textsc{Normal} exploration. The \textsc{PRR} mode is entered only from \textsc{Normal} and only when $\mathit{unimp\_num}$ exceeds $\theta_{\mathrm{PRR}}$. 
This ensures that the PRR search is triggered when persistent non-improvement indicates that a deeper dive is needed (lines 25-26).
After the mode decisions, the algorithm removes a vertex $r$ from $C$ using \textsc{SelectRemove} (line~27), and then chooses a vertex $x$ to add back according to the current mode: \textsc{PRRSelect} (Algorithm \ref{alg:prr}) for \textsc{PRR}, \textsc{NormalAdd} for \textsc{Normal}, or \textsc{RepairAdd} for \textsc{Repair} (lines~28--36). 
After each insertion, $\mathit{q\_gain}$ is increased by  $\bigl|\bigl(Q(C\cup\{x\})\setminus Q(C)\bigr)\setminus\{x\}\bigr|$ (line~35). 
If the mode is \textsc{Normal},  $Z(C) = \emptyset$, and $\mathit{q\_gain}$ has reached $\theta_{\mathrm{ERE}}$, the \textsc{ERESelect} (Algorithm \ref{alg:ere}) performs an additional reduction on $C$, and $\mathit{q\_gain}$ is reduced by $\theta_{\mathrm{ERE}}$ (lines~37--38). 
Finally, $\mathit{unimp\_num}$ increments by 1 (line~39), and the loop proceeds. 
When time runs out, the algorithm returns the best VC $C^*$ found (line~40).


Each local-search step inserts or removes a constant number of vertices. Updating the auxiliary data structures requires only inspecting the closed neighborhoods of the removed vertex $r$ and the added vertex $x$, which takes $O(d(r)+d(x))=O(|V|)$ time. The \textsc{RepairAdd} procedure enumerates at most $\sqrt{|E|}=O(|V|)$ uncovered edges. The \textsc{ERESelect} and \textsc{PRRSelect} procedures examine only a fixed number of candidate vertices per invocation and take $O(|V|)$ time. \textsc{FoldOpt} requires $O(|V|^3)$ time in the worst case. Let $T$ be the total number of iterations. The local search therefore requires $O(T|V|^3)$ time. Additionally, the initial construction by \textsc{GCNInitVC} costs $O(|V||E|\log |E|)$. Combining these two parts gives an overall complexity of $O(|V||E|\log |E|+T|V|^3)$.

\subsection{Two Optimization Strategies}
 \textsc{GRALS} uses two strategies, ERE and PRR, to exploit redundancy for enhancing the search process. ERE is an explicit structural exchange that expands the search neighborhood during infeasible searches. 
Unlike standard repair methods, which simply insert an endpoint of an uncovered edge, \textsc{ERESelect} (Algorithm \ref{alg:ere}) samples $\theta_{\mathrm{ERE}}$ vertices from the set $V\setminus C$ and evaluates them based on their capabilities of releasing redundancy, irrespective of their incidence to uncovered edges (line~1). It then collects sampled vertices with $R_C(v)\geq2$ into the set $\mathcal{X}_{\mathrm{MR}}$ (line 2). If $\mathcal{X}_{\mathrm{MR}}\neq\varnothing$, a vertex with the maximum $R_C$ is selected, with ties broken by larger $cn_C$ (lines~3--4). If no such vertices exist, a vertex $x$ is selected that satisfies $R_C(x)=1$ and $cn_C(x)>2cn_C(u_x)$, where $u_x$ is the unique vertex in $S_C(x)$. The algorithm returns $C$ without an exchange if no suitable $x$ is found (lines~5--8). Finally, $x$ is inserted into $C$, and a vertex $u\in S_C(x)$ is randomly removed from $C$ (lines~9--11).

{
\setlength{\textfloatsep}{4pt}
\setlength{\floatsep}{4pt}
\begin{algorithm}[!t]
\scriptsize

\SetAlgoSkip{smallskip}
\SetNlSkip{1pt}
\setlength{\algomargin}{0em}
\setlength{\rightskip}{-0.6em}
\SetInd{0.2em}{0.45em}
\caption{\textsc{ERESelect}}
\label{alg:ere}
\KwIn{A candidate set $C$ with $Z(C)=\varnothing$, a budget $\theta_{\mathrm{ERE}}$}
\KwOut{Updated candidate set $C$}

Sample $\theta_{\mathrm{ERE}}$ vertices outside $C$\;
$\mathcal{X}_{\mathrm{MR}}\leftarrow\{v\mid R_C(v)\geq 2\}$ among the sampled vertices\;

\eIf{$\mathcal{X}_{\mathrm{MR}}\neq\varnothing$}{
    $x\leftarrow$ vertex in $\mathcal{X}_{\mathrm{MR}}$ with maximum $R_C$.
    Ties broken by larger $cn_C$\;
}{
    Select a vertex $x$ from $V\setminus C$ satisfying $R_C(x)=1$ and $cn_C(x)>2\,cn_C(u_x)$\;
    \If{\textup{no such $x$ exists}} {\Return{$C$}\;}
}

$u\leftarrow$ a random vertex in $S_C(x)$\;
$C\leftarrow (C\cup\{x\})\setminus\{u\}$\;
\Return{$C$}\;

\end{algorithm}
}


PRR modifies only the repair-selection rule within the standard break-and-repair process, without altering its overall structure. When the search stagnation ($\mathit{unimp\_num}>\theta_{\mathrm{PRR}}$), \textsc{PRRSelect} (Algorithm \ref{alg:prr}) replaces \textsc{NormalAdd} and iteratively samples at most $k$ uncovered edges from $U(C)$ (line~1); for each sampled edge $e=uv$, it examines its endpoints $y\in\{u,v\}$ and collects those satisfying $y\neq r$, $\mathit{conf\_change}(y)=1$, and $R_C(y)>0$ into a set $A$ (lines~2--3). If $A\neq\emptyset$, it returns the vertex in $A$ with maximum $R_C$, breaking ties by larger $\mathit{dscore}$ and then by smaller $\mathit{time\_stamp}$ (lines~4--6); if no such endpoint is found after $k$ attempts, it falls back to \textsc{NormalAdd} (line 7).

\subsection{Optimization Guarantee}
Given an input instance $G = (V, E)$, let \( C^* \) represent the current best VC. During the search for infeasible solutions, GRALS maintains a candidate set \( C \) with \( |C| = |C^*| - 1 \). The goal of ERE is to minimize \( |U(C)| \) without increasing \( |C| \). 
ERE follows two routes: Route I (lines 2-4 in Algorithm \ref{alg:ere}) aims to produce additional zero-loss vertices. By deleting such a vertex and adding an endpoint of an uncovered edge, \( |C| \) is preserved while \( U(C) \) is reduced. Route II (lines 6-8 in Algorithm \ref{alg:ere}) preserves or enhances the double-cover structure when only one critical vertex can be released. In the following discussion,  we denote by \( x (\in V \setminus C) \) the vertex selected by ERE for addition to \( C \), and \( u (\in C) \) the vertex selected for removal from \( C \). We define \( C' = (C \setminus \{u\}) \cup \{x\} \).

\textbf{Proposition 1}
$|C'|=|C|$ and $U(C')\subseteq U(C)$. The inclusion is strict if $dscore(x)\geq 1$. 

\textit{Proof.}
$|C'|=|C|$ is obvious. Since $u\in S_C(x)$, $u$ is critical and $x$ is the unique neighbor of $u$ outside $C$. Thus, inserting $x$ makes $u$ zero-loss, and removing $u$ results in no uncovered edge. So, $U(C')\subseteq U(C)$. When $dscore(x)\geq 1$, adding $x$ to $C$  ensures that at least one previously uncovered edge becomes covered, leading to \( U(C') \subset U(C) \). \hfill \( \square \)

Proposition 1 supports the exchange in Algorithm~\ref{alg:ere} (lines 9–10). In particular, 
if \( x \) is incident with an uncovered edge, it enables immediate progress in the repair process.

{
\setlength{\textfloatsep}{3pt}
\setlength{\floatsep}{3pt}
\setlength{\abovecaptionskip}{3pt}
\begin{algorithm}[!t]
\scriptsize

\SetAlgoSkip{smallskip}
\SetNlSkip{1pt}
\setlength{\algomargin}{0em}
\setlength{\rightskip}{-0.6em}
\SetInd{0.2em}{0.45em}
\caption{\textsc{PRRSelect}}
\label{alg:prr}
\KwIn{A candidate set $C$, a removed vertex $r$, budget $k$}
\KwOut{A repair vertex $x$}

\For{$i\leftarrow 1$ \KwTo $k$}{
    $e=uv\leftarrow$ an edge selected randomly from $U(C)$\;
    $A\leftarrow \{y\in\{u,v\}\mid y\neq r \land \mathit{conf\_change}(y)=1 \land R_C(y)>0\}$\;
    \If{$A\neq\emptyset$}{
        $x\leftarrow$ vertex in $A$ with maximum $R_C$.
        Ties broken by larger $\mathit{dscore}$, then smaller $\mathit{time\_stamp}(x)$;$no\_gain\_steps\leftarrow0$\;
        \Return{$x$}\;
    }
}
$x\leftarrow\textsc{NormalAdd}(C), no\_gain\_steps \leftarrow no\_gain\_steps + 1$\;
\Return{$x$}\;
\end{algorithm}
}

\textbf{Proposition 2}
Suppose that $x$ is selected through Route I. Let $w \in S_C(x)\setminus \{u\}$ be a vertex such that $uw\notin E$, and let $y$ be a vertex that is incident with an edge in $U(C')$. Then, it follows that $U((C'\setminus \{w\})\cup \{y\}) \subset U(C')$. 

\textit{Proof.}
Since  $w\in S_C(x)$, $x$ is the unique neighbor of $w$ outside $C$. Thus, $\ell_{C'}(w)=0$ because $uw\notin E$. Therefore, removing \( w \) from \( C' \) does not create any uncovered edges. Furthermore, adding \( y \) to \( C' \) covers at least one previously uncovered edge and does not create any new uncovered edges. Thus, the conclusion holds.  \hfill \( \square \)

Proposition~2 implies that by adding \( x \) to \( C \), a set \( A \) can be formed containing multiple zero-loss vertices. However, the ERE removes only one of these vertices, denoted as \( u \). While the elimination of \( u \) may result in some vertices within \( A \setminus \{u\} \) no longer being zero-loss, those vertices in \( A \setminus \{u\} \) that are not adjacent to \( u \) still remain strong candidates for safe removal in subsequent iterations.

Note that removing all $k$ zero-loss vertices simultaneously after adding a new vertex would reduce the size of the candidate set to $|C| + 1 - k$. For $k \geq 2$, this could potentially bring the search below a cardinality level where a feasible VC exists. Therefore, ERE only performs one immediate deletion of a vertex at a time, and the remaining redundancy is addressed through subsequent remove-add iterations.

\textbf{Proposition 3}  
Let \( x \) be selected through Route II, and let \( S_C(x) = \{u\} \). Then, we have:  
$
D(C') - D(C) = cn_C(x) - cn_C(u) - 1,
$
where \( D(X) \) denotes the number of double-covered edges by the set \( X \).

\textit{Proof.}
When \( x \) is added to \( C \), it generates \( cn_C(x) \) double-covered edges. Conversely, removing \( u \) from \( C \) eliminates \( cn_C(u) \) existing double-covered edges, along with one newly created double-covered edge \( ux \). Hence, we can conclude that  
$D(C')-D(C)=cn_C(x)-cn_C(u)-1$.  \hfill $\square$

More generally, consider \( \lambda \geq 1 \) as a positive integer such that \( cn_C(x) > \lambda cn_C(u) \). Under this condition, Proposition 3 indicates that \( D(C') - D(C) \geq (\lambda - 1) cn_C(u) \). For \( \lambda = 1 \), there is no increase in the number of double-covered edges. When \( \lambda \geq 2 \) and \( cn_C(u) \geq 1 \), the number of double-covered edges strictly increases. In our experiments, we choose the smallest value of \( \lambda \) that allows for an increase in double-covered edges, i.e., $\lambda=2$.

Regarding PRR, any vertex \( x \) that is returned prior to the fallback is an uncovered-edge endpoint with \( R_C(x) > 0 \) (see Algorithm~\ref{alg:prr}, lines~1–6). Therefore, for a vertex \( u \in S_C(x) \), Proposition~1 indicates that adding \( x \) to the candidate set \( C \) while removing \( u \) from \( C \) maintains the size of the candidate set and reduces the uncovered-edge set.

\section{Experimental Evaluation} \label{section-4}
We evaluate GRALS on a broad set of large-scale MVC benchmarks and investigate the impact of its key components. 

  \subsection{Experimental Setup}

We compare GRALS with five state-of-the-art heuristic algorithms for MVC, including 
FastVC \cite{Fan2015}, NoiseVC \cite{MaFSLS16}, EAVC \cite{quan2021local}, TIVC \cite{Zhang2023}, and InfVC \cite{Sun2024}. The source code for these algorithms was kindly provided by their authors.

Following recent MVC heuristic studies, we use a unified time
  limit of 1000 seconds for all algorithms. Each instance is run
  10 times with random seeds from 1 to 10. All algorithms are implemented in C++
  and compiled with g++ -O3. For each baseline, we adopt the default
  parameter settings reported in the original paper. All experiments are conducted on a server with two Intel(R) Xeon(R) Platinum 8362 CPUs at 2.80 GHz, Tesla V100 32GB GPU, and 512 GB of
  RAM.

Our benchmark for testing consists of 346 graph instances primarily sourced from the Network Data Repository \cite{rossi2015network}, in accordance with recent MVC studies. These instances are used exclusively for evaluation and are entirely distinct from the graphs used to train the GCN.

\subsection{Results on All Instances}

Table~\ref{tab:pairwise_compare} presents a summary of the overall results. The ``Best'' indicates the number of instances on which the best solution achieved by an algorithm over 10 runs matches the best-known solution. The ``Mean'' denotes the number of instances where a solver achieves the smallest mean VC size over 10 runs, among all algorithms whose best results meet the best-known solution.
In the ``GRALS'' row, each entry formatted as ``wins-losses'' specifies the number of instances where GRALS outperforms versus underperforms the respective baseline; ties have been excluded from this account. GRALS records more wins than losses against each baseline. Even when compared to the leading algorithm InfVC, GRALS achieves an impressive record of 42 wins and 10 losses. 


{
\setlength{\textfloatsep}{4pt}
\setlength{\floatsep}{4pt}
\setlength{\abovecaptionskip}{4pt} 
\begin{table}[!t]
\centering

\scriptsize
\setlength{\tabcolsep}{5pt}
\begin{tabular}{lcccccc}
\toprule
& EAVC & FastVC & GRALS & InfVC & NoiseVC & TIVC \\
\midrule
Best 
& 206/346 & 204/346 & \textbf{335/346}
& 302/346 & 217/346 & 198/346 \\
\midrule
Mean 
& 186/346 & 187/346 & \textbf{330/346}
& 291/346 & 183/346 & 171/346 \\
\midrule
GRALS
& 137-4 & 140-3 & --
& 42-10 & 128-5 & 143-4 \\
\bottomrule
\end{tabular}
\caption{Overall performance comparison on 346 instances.}
\label{tab:pairwise_compare}
\vspace{4pt}
\end{table}
}

{
\setlength{\textfloatsep}{3pt}
\setlength{\floatsep}{3pt}
\setlength{\abovecaptionskip}{3pt}
\begin{figure*}[!t]
  \centering
  \includegraphics[width=0.9\textwidth]{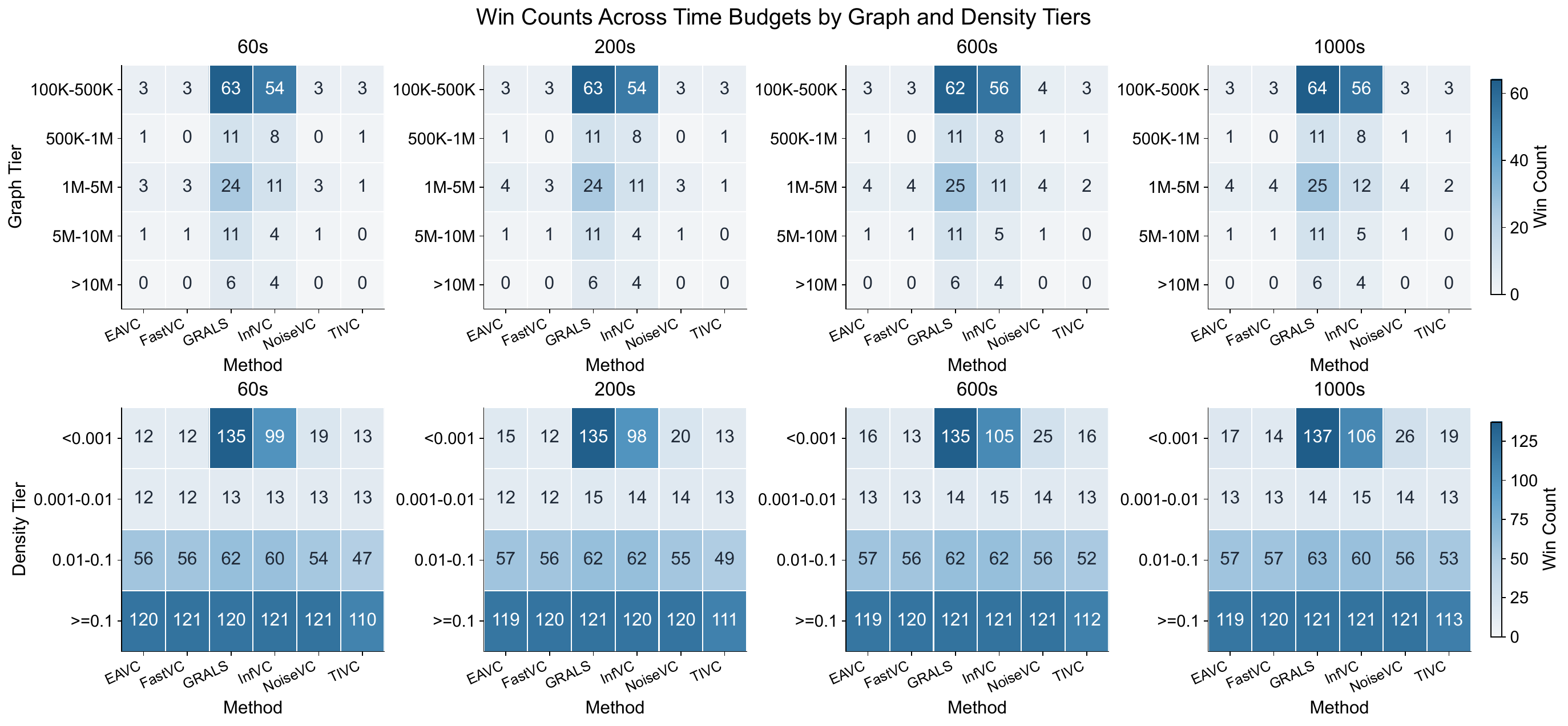}
  \caption{Win counts across four time budgets by graph scale and edge density.}
  \label{fig:win_count_heatmap}
\end{figure*}
}

Figure~\ref{fig:win_count_heatmap} displays the win counts recorded at 60, 200, 600, and 1000 seconds. GRALS outperforms all other methods across all graph-scale tiers and achieves the highest win counts across most density tiers. To illustrate the progression of GRALS's performance over time, Figure~\ref{fig:gcnlp_vs_infvc} highlights the 12 massive instances on which GRALS exhibits the most significant VC-size improvements over InfVC at the 1000-second mark. GRALS establishes its lead early and sustains this advantage throughout the run duration.


{
\setlength{\textfloatsep}{3pt}
\setlength{\floatsep}{3pt}
\setlength{\abovecaptionskip}{3pt}
\begin{figure*}[!t]
  \centering
  \includegraphics[width=0.98\textwidth]{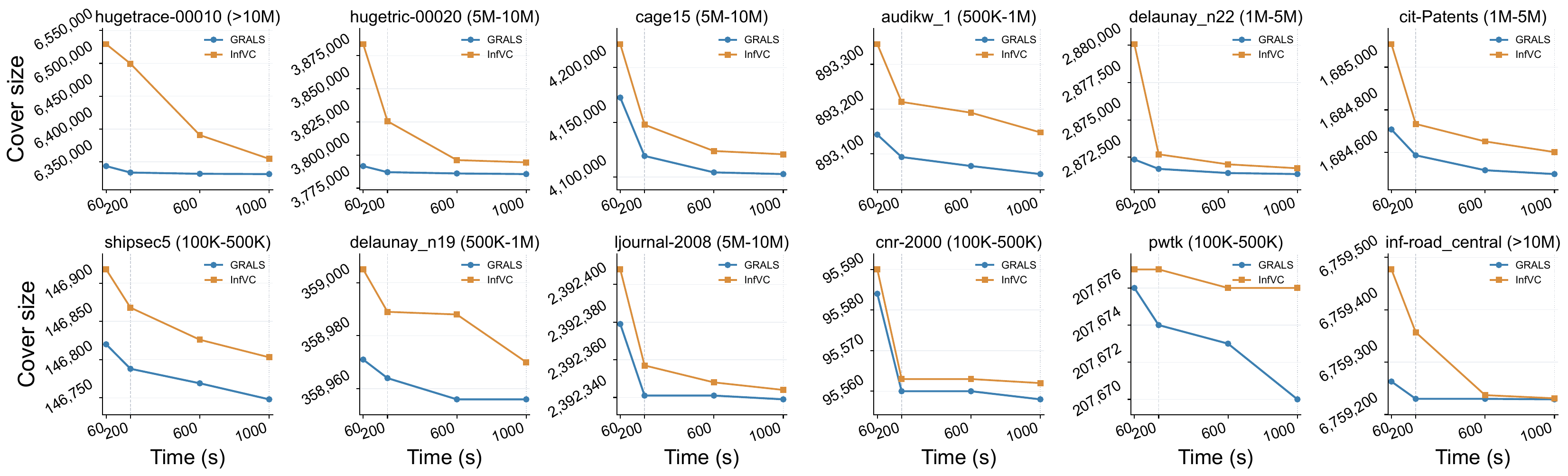}
  \caption{Anytime comparison of GRALS and InfVC on 12 selected large instances.}
  \label{fig:gcnlp_vs_infvc}
\end{figure*}
}


{
\setlength{\textfloatsep}{4pt}
\setlength{\floatsep}{4pt}
\setlength{\abovecaptionskip}{4pt}
\begin{table}[!t]

\centering
\scriptsize
\setlength{\tabcolsep}{6pt}

\begin{tabular}{lrrrrrr}
\toprule
Range & Total & InfVC & GCNInitVC & Eq & $\min\Delta$ & $\max\Delta$ \\
\midrule
\multicolumn{7}{l}{Vertex Scale} \\
\midrule
$<$1K       & 122 & 29 & 36 & 57 & $\phantom{+}-5$    & $+5$ \\
1K--10K     & 78  & 18 & 32 & 28 & $\phantom{+}-4$    & $+102$ \\
10K--100K   & 23  & 3  & 15 & 5  & $\phantom{+}-10$   & $+1054$ \\
100K--1M    & 80  & 15 & 55 & 10 & $-133$  & $+7627$ \\
$\ge 1$M    & 43  & 9  & 28 & 6  & $-2381$ & $+190469$ \\
\midrule
\multicolumn{7}{l}{Edge Density} \\
\midrule
$<0.001$    & 146 & 27 & 98 & 21 & $-2381$ & $+190469$ \\
0.001--0.01 & 16  & 2  & 8  & 6  & $\phantom{+}-4$    & $+102$ \\
0.01--0.1   & 63  & 21 & 29 & 13 & $\phantom{+}-4$    & $+131$ \\
$\ge 0.1$   & 121 & 24 & 31 & 66 & $\phantom{+}-5$    & $\phantom{+}+4$ \\
\bottomrule
\end{tabular}
\caption{Initial VC comparison based on graph scale and edge density.}
\label{tab:perf_stat}
\vspace{1pt}
\end{table}
}

\subsection{Effect of GCN-Guided Initialization}
Table~\ref{tab:perf_stat} presents a comparison of the initial VCs produced by InfVC and GCNInitVC. We use $\Delta$ to denote the difference between the cardinalities of VCs generated by InfVC and GCNInitVC.  A positive value indicates that GCNInitVC yields a smaller VC. Overall, GCNInitVC demonstrates superior performance on 166 instances, falls short on 74 instances, and results in ties on 106 instances. Its advantages are particularly prominent with large sparse graphs. For instances containing at least 100K vertices, GCNInitVC secures 83 wins against 24 losses, achieving a maximum improvement of 190,469. For instances with densities below 0.001, it records 98 wins and 27 losses. However, on dense graphs, both methods perform comparably, with 66 ties and $\Delta$ values ranging from -5 to 4. This similarity may be attributed to the dense structure: overlapping neighborhoods make vertex features highly similar, diminishing the discriminative power of probability.

Additionally, we compare \textsc{GCNInitVC} with a random-probability variant \textsc{RanPro\_Init} (see Table~S1 in the supplementary material). \textsc{GCNInitVC} outperforms \textsc{RanPro\_Init} on 223 instances and falls short on 34. These findings indicate that the learned structural prior provides advantages in initializing large sparse graphs.

\subsection{Ablation Study}


{
\setlength{\textfloatsep}{4pt}
\setlength{\floatsep}{4pt}
\setlength{\abovecaptionskip}{4pt}
\begin{table}[!t]
\centering

\scriptsize
\setlength{\tabcolsep}{8.5pt}

\begin{tabular}{lrrr@{\hspace{7pt}}|lrrr}
\toprule
Variant & W & L & T & Variant & W & L & T \\
\midrule
w/o GCNInit & 40 & 8  & 298 &
w/o ERE     & 34 & 8  & 304 \\

InfVC\_GCN  & 41 & 4  & 301 &
w/o PRR     & 25 & 12 & 309 \\
\bottomrule
\end{tabular}
\caption{Component ablation on all 346 test instances. 
W, L, and T denote wins, losses, and ties, respectively.}
\label{tab:ablation}
\vspace{3pt}
\end{table}
}

We evaluated GCNInit by comparing it to the w/o GCNInit variant, which utilizes the original InfVC initialization while maintaining the  GRALS local search. As shown in Table~\ref{tab:ablation}, GRALS achieved 40 wins and 8 losses against this variant.
To further isolate the contribution of GCN-guided initialization from that of redundancy-aware local search, we created a new variant InfVC\_GCN, which uses the same initialization as GRALS while retaining the InfVC local search. The results demonstrate that GRALS achieved 41 wins and 4 losses against InfVC\_GCN, suggesting that its benefits extend beyond the initialization process alone. Additionally, we examined the individual impacts of the ERE and PRR components by disabling them one at a time. GRALS recorded 34 wins and 8 losses against the variant without ERE and 25 wins and 12 losses against the variant without PRR. These results provide empirical support for the assertion that both mechanisms enhance the search process.

{
\setlength{\textfloatsep}{3pt}
\setlength{\floatsep}{4pt}
\setlength{\abovecaptionskip}{4pt}
\begin{figure*}[!t]
\centering
\includegraphics[width=0.98\textwidth]{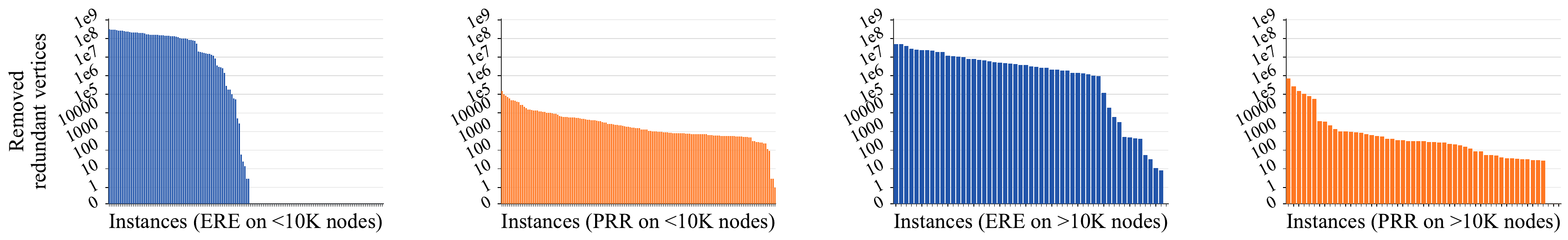}
\caption{Successful redundancy-induction events generated by ERE and PRR.}
\label{fig:ere_prr_compare}
\end{figure*}
}



To examine the functionality of the ERE and PRR strategies, we record their
successful redundancy-induction events.
An event is considered \textit{successful} when a mechanism creates a zero-loss vertex that is
subsequently removed safely. Figure~\ref{fig:ere_prr_compare} illustrates the maximum number of successful events across 10 runs for each instance. 

Both mechanisms effectively create exploitable redundancy during the search process. ERE generates a higher number of events, whereas PRR produces fewer events as it is triggered only during periods of stagnation. The event counts confirm that both mechanisms function as intended. These findings indicate that ERE actively fosters redundancy throughout the standard search process, while PRR offers valuable support during moments of stagnation.

\subsection{Additional Experiments}
{\color{black} Due to space limitations, additional experiments are reported in the supplementary material (Tables S2--S3). They include redundancy-aware extensions for minimum dominating set (MDS) and set cover problems, parameter sensitivity analyses, and experimental settings. The MDS variant achieves 27 wins and 6 losses against NuMDS~\cite{sun2025numds}. The set cover variant achieves 3 wins and 1 loss against NuSC~\cite{luo2022nusc}.}

\section{Conclusion} \label{section-5}

We present \textsc{GRALS}, a GCN-guided local search framework with redundancy-aware mechanisms (ERE and PRR) for MVC. Evaluated on 346 benchmark instances, \textsc{GRALS} outperforms state-of-the-art MVC algorithms. Ablation studies confirm the distinct contributions of the learned initialization, redundancy guidance, ERE, and PRR. Behavioral analysis reveals that ERE is critical during regular search, while PRR acts as a complementary strategy when the search stagnates. 
Future work includes integrating redundancy-aware structural information into GCNs to enhance vertex evaluation and generate informative priors for local search. We will model neighborhood overlap in dense graphs to better identify and exploit potential redundancy.

\bigskip

\bibliography{aaai2027}


\end{document}